\documentclass[letterpaper,conference]{IEEEtran}

\IEEEoverridecommandlockouts

\usepackage{amsmath,amssymb,amsfonts}
\usepackage{algorithmic}
\usepackage{graphicx}
\usepackage{textcomp}
\usepackage{xcolor}
\usepackage{tabularx} 
\usepackage{array}
\usepackage{pdfcomment}
\usepackage{soul}
\usepackage{float}

\usepackage{hyperref}
\def\BibTeX{{\rm B\kern-.05em{\sc i\kern-.025em b}\kern-.08em
T\kern-.1667em\lower.7ex\hbox{E}\kern-.125emX}}

\begin{document}

\title{Guiding Empowerment Model: Liberating Neurodiversity in Online Higher Education\\}

\author{
\begin{minipage}[t]{0.25\textwidth}
\centering
Hannah Beaux \\
\textit{Program Experience Strategy, WGU} \\
Portland, USA \\
hannah.beaux@wgu.edu
\end{minipage}
\hfill
\begin{minipage}[t]{0.24\textwidth}
\centering
Pegah Karimi \\
\textit{Program Experience Strategy, WGU} \\
Indianapolis, USA \\
pegah.karimi@wgu.edu
\end{minipage}
\hfill
\begin{minipage}[t]{0.25\textwidth}
\centering
Otilia Pop \\
\textit{Program Experience Strategy, WGU} \\
San Francisco, USA \\
otilia.pop@wgu.edu
\end{minipage}
\hfill
\begin{minipage}[t]{0.25\textwidth}
\centering
Rob Clark \\
\textit{Program Experience Strategy, WGU} \\
Salt Lake City, USA \\
rob.clark@wgu.edu
\end{minipage}
}

\maketitle
\begin{abstract}
In this innovative practice full paper, we address the equity gap for neurodivergent and situationally limited engineering or computing learners by identifying the spectrum of factors that impact learning and acknowledging the fluctuations of learner function. Educators have shown a growing interest in identifying learners' cognitive abilities and learning preferences to measure their impact on academic achievement. These needs, however, are often addressed via one-size-fits-all approaches leaving the burden on disabled students to self-advocate or tolerate inadequately conducive conditions for their learning. Emerging frameworks guide the support of a neurodiverse learner population in curriculum and assessment activities through instructional approaches, such as online education. However, the application of these frameworks is disaggregated, and the technology interventions recommended for the online learning environment remain insubstantially supportive resulting in disparity, particularly for those with undisclosed learning or developmental disabilities and situational limitations. In this article, we integrate a neurodivergent perspective through secondary research of around 100 articles to introduce a comprehensive Guiding Empowerment Model involving key cognitive and situational factors that contextualize day-to-day experiences affecting learner ability. We illustrate the model by synthesizing common formations of these factors to facilitate three sample student profiles that highlight fluctuating user perceptions and explore initially evident actionable user problems in functioning. We use this model to evaluate sample learning platform features and other supportive technology solutions that potentially address the needs of the neurodiverse learner population represented in the learner profiles. The proposed approach augments frameworks such as Universal Design for Learning to consider factors including various sensory processing differences, social connection challenges, and environmental limitations. We suggest that by applying the model through technology-enabled features such as customizable task management, guided varied content access, and guided multi-modal collaboration, major learning barriers of neurodivergent and situationally limited learners will be removed to activate the successful pursuit of their academic goals.
\end{abstract}


\section{Introduction}
Neurodivergence \cite{b1, b2} is becoming an increasingly necessary consideration when designing learning experiences. Historically, support for neurodivergent students has been referred to student disability offices in higher education institutions that require official diagnoses before support is rendered \cite{b2, b3}. While often withheld from those without a documented diagnosis, accommodations can be ineffective \cite{b4} or irrelevant even for those with the proper documentation, resulting in reduced help-seeking behaviors in at-risk neurodivergent students \cite{b5, b6}.

Prior work has expounded on the disparity neurodivergent students face when pursuing STEM education \cite{b7, b8} in particular, and solutions have risen, such as online learning \cite{b9} which offers flexibility for learning time and place, cognitive styles \cite{b10} that involve the various ways learners process information, and Universal Design for Learning \cite{b11}, which aims to increase access and participation in meaningful learning experiences. However, there is a gap in describing effective implementation for these largely content-focused, instructor-led instructional approaches with learners who require more personalized or adaptive support in an online asynchronous environment. 

The authors seek to address this inequity through the synthetization and application of a Guiding Empowerment Model. The model is based on secondary research of evolving educational theories and practices and is comprised of 31 scaled environmental and cognitive factors that impact meaningful learner interaction and progress. This practice endeavors to achieve several goals by addressing various dimensions. Firstly, it applies an underlying conceptual model to identify and justify the essential needs of neurodiverse learners in an asynchronous online competency-based learning environment. Secondly, the practice explores these needs through the lens of three distinct sample personas, providing insight into their diverse requirements. Lastly, the practice outlines a process that detects learner needs using analytical software, implements effective support systems, and facilitates timely accommodations within the learning platform via digitally surfaced resource recommendations to increase momentum, particularly for learners exhibiting characteristics that resemble neurodivergent traits. A variety of exploratory technical solutions are introduced as components of the learning support system. This article addresses the contextual barriers neurodivergent STEM learners face in their academic pursuits by proposing strategies for deliberate and active organizational commitment to the equitable academic achievement of a neurodiverse learner population.

\section{Relevant literature and existing practices}
\subsection{Neurodiversity}
Neurodiversity, which describes the spectrum of cognition and behavior across both neurotypical and neurodivergent individuals \cite{b1}, \cite{b2}, is a key factor in learner efficacy. While some view neurodivergence as a deviation from typical cognitive processing and environmental response \cite{b12}, others, particularly within the neurodivergent community, argue these differences are natural variations along the Human Spectrum \cite{b13}. Nonetheless, neurodivergent learners often struggle to succeed in conventional higher education settings despite accommodations and face a higher risk of burnout and other mental health effects \cite{b3}.
\subsection{Online learning approaches}
Online learning at the college level has become pervasive \cite{b14} and, more recently, synonymous with education, primarily due to constant technological advancements. No-code online authoring and delivery solutions \cite{b9} enabled transformative, scalable, and equitable learning approaches \cite{b15, b16}. One such approach, competency-based education \cite{b17, b18}, emphasizes the practical application of acquired skills and offers greater flexibility and adaptivity \cite{b19} to diverse learner populations \cite{b20} via self-directed, asynchronous online learning \cite{b21}.
\subsection{Cognitive Styles}
Not to be confused with learning styles that sort learners into static categories of instructional modes \cite{b22}, cognitive styles distinguish three key cognitive spectra: field dependence or independence, which refers to learners’ tendency to think about the big picture versus small details, holistic or analytical thinking, and reflective or impulsive thinking \cite{b10}. This approach has been used to support students in traditional instructor-led  classrooms and recently in a case of chatbot-assisted content delivery \cite{b23} but has not been widely applied in an asynchronous online competency-based university setting for computing or engineering students.
\subsection{Neuro-inclusive Instructional Approaches}
Often, learners without a documented disability who experience cognitive or environmental challenges \cite{b24, b25} are left to manage their challenges on their own or via inadequate support \cite{b5}. Additionally, those eligible for accommodations face lingering challenges \cite{b4} in the university setting. Two instructional approaches that incorporate inclusivity and learner well-being needs have emerged as transformative to educational equity: {trauma-informed} instruction \cite{b26}
and Universal Design for Learning (UDL) \cite{b11, b27, b28}.
Trauma can manifest in a variety of behaviors that impact academic achievement, such as social avoidance, impulsivity, distractibility, inflexibility, and risk-taking challenges \cite{b29}. Trauma-informed instruction acknowledges educators' responsibility to disrupt the systems enabling trauma by structuring new, equitable learning experiences that consider learners' holistic needs. Educators cannot distinguish between learners with non-apparent disability or trauma and those without these circumstances \cite{b30}, so a human-centered lens must shape their instructional approach.

The Universal Design for Learning (UDL) framework, which is grounded in extensive literature on the educational needs of neurodivergent students embodies inclusive learning design \cite{b11} by providing multiple means of representation, engagement, and expression to accommodate the variability of a diverse learner population. Some key principles of UDL are providing multiple formats of instructional materials, such as both auditory and visual information, facilitating learner development of executive function skills through agency\cite{b31} and developing learners' self-awareness and regulation strategies.
While growing evidence documents the positive effects of implementing trauma-informed instruction and UDL on learner achievement \cite{b32}, and specific strategies exist for supporting engineering students through UDL 
\cite{b33}, these frameworks are based on a synchronous, instructor-led environment and provide limited direction for addressing a nuanced and complex adult online learner population who may face a variety of distinct socioeconomic, neurocognitive, and social factors that can impact learner progress \cite{b6}, \cite{b24}. There is also a gap in guiding the prioritization of impactful academic technology solutions to support learners based on their fluctuating needs.
In previous work, sensory differences \cite{b34, b35, b36} in particular have been associated with varying levels of impulsivity, rejection sensitivity, overactivity, multitasking \cite{b37}, demand nonconformity 
\cite{b38}, and adaptability \cite{b39}. Sensory differences can also affect individuals’ isolation from social support \cite{b40}. For example, they may experience challenges in social interaction, emotional regulation, and help-seeking or teamwork behaviors required for academic performance and be more likely to drop out of educational programs \cite{b41}. Recurrent sensory challenges and discomfort may also cause anxiety, stress, and avoidance \cite{b42} or avoidant-perceived behaviors \cite{b6} impacting learner well-being and academic performance \cite{b41}. Learners’ ideal environments may vary from day-to-day, so educators must explore interventions that both give learners agency and support to modify their environments to suit their own fluctuating sensory needs \cite{b11,b31,b43} and anticipate those fluctuations to adapt instruction automatically \cite{b19}.
\section{Guiding Empowerment Model (GEM)}
Guiding Empowerment Model (GEM) is a conceptual framework that compiles secondary research to highlight the day-to-day cognitive and environmental factors that contextualize \cite{b44} learner functioning and behavior and identify actionable problems mitigable through technical solutions. Apparent disability, non-apparent disability, and situational limitations can all fluctuate along a spectrum, causing variations in the way online learners interact with their environment, other people, and their course materials. GEM identifies 31 dynamic characteristics described in the literature as essential to how learners process new information. It can be used to detect and address learner needs by applying UDL and trauma-informed practices through technological supports such as AI-guided learner choice \cite{b23}. Table I categorizes these 31 scales into six interrelated groups: time and energy, sensory processing, social cognition, reasoning, executive function, and core skills. 
\begin{table}[H]
\centering
\caption{The six categories and 31 scales of Guiding Empowerment Model}
\label{tab:single-column}
\begin{tabularx}{\linewidth}{|X|}
\hline
\textbf{GUIDING EMPOWERMENT MODEL}  \\[2pt] 
\hline
\textbf{Time and Energy}

Physical engagement. Mental engagement.  \\
\hline
\textbf{Sensory Processing} 

Visual seeking (high to low), Visual avoiding, (high to low) Auditory seeking (high to low), Auditory avoiding (high to low) Tactile seeking (high to low), Tactile avoiding (high to low). \\
\hline

\textbf{Social Cognition}

Rejection reactive to rejection persistent, Demand conforming to demand avoiding, Social connection desire to social avoidance,  Empathy to apathy, Social stability to social
flexibility. \\
\hline

\textbf{Reasoning}

Adaptivity to rigidity, Emotional thinking to critical thinking, Decisive to indecisive, Less risk comfort to more risk comfort, Reliability to impulsivity. \\
\hline

\textbf{Executive Function}

Time management ability, Hyperfocused to distractible, Multitasker to single tasker, Memory, Repetitive tasking to varied tasking, Big picture oriented to detail oriented. \\

\hline
\textbf{Core Skills}

Subject matter efficacy, Written language, Spoken language, Technology efficacy, Self awareness. \\
\hline
\end{tabularx}
\end{table}
\subsection{Time and Energy}
This category pertains to how learners expend their time and energy throughout their day. Various day-to-day environmental and internal factors can influence the engagement and momentum of online learners \cite{b45}. While learners in an online cohort are all assigned the same duration to complete their courses, they may have varied actual time and energy \cite{b46} to spend on study-related activities due to other commitments of daily living \cite{b47} and the physical environments where learners spend their days may imply varied situational needs. Factors such as interruptions, distractions, or fatigue \cite{b48} can affect how productive a learner is in progressing toward their academic goals \cite{b49}.
Learner life, in terms of time and energy represents physical engagement and mental engagement. Some tasks require both physical and mental exertion, such as caregiving. Some require high mental engagement but low physical engagement, such as job-related web-calls and some require physically demanding but less mentally taxing tasks such as driving or running. A learner’s free time, a potential opportunity for study, can be impacted not only by time spent on these other tasks leading to fatigue \cite{b38}, but also by the physical environment surrounding their free time, whether it is conducive to focus or fogged up by distracting sensory stimuli. While educators may struggle to mitigate the environmental barriers in online learners’ study spaces, this article suggests several tools and policies that can support learners challenged by distracting environments of study.
\subsection{Sensory Processing}
Sensory processing refers to the ways \cite{b49} the nervous system receives and interprets sensory information from the environment \cite{b50} to enable meaningful interaction with the world \cite{b51,b52}. Differences \cite{b35, b53} in sensory processing can affect behaviors \cite{b44, b54, b55} at times impairing learner ability \cite{b56}. Sensory processing theory describes four distinct patterns of response to sensory information in the environment: sensory avoiding, a tendency to actively avoid or experience discomfort with certain sensory stimuli \cite{b55}; sensory seeking, a tendency to seek out intense sensory experiences; sensory sensitive, a heightened reaction to sensory stimuli; and sensory unregistered, a lessened awareness or minimal response to sensory input \cite{b54,b57}. An individual’s sensory orientation can vary based on situational or environmental factors and may fluctuate over time \cite{b44,b58} without their awareness to its impact on their ability to adhere to the norms \cite{b38} of academic pursuit \cite{b20}. GEM defines the Sensory Processing category as learners’ visual, tactile, and auditory reactivity patterns which directly \cite{b41} and indirectly \cite{b12,b38,b39,b40,b42,b59} relate to learner retention in academic programs.
\subsection{Social Cognition}
Appropriately applied social interaction in learning environments is positively associated with academic achievement \cite{b60,b61,b62,b63}. The Social Cognition category is defined by five scales: rejection reactivity, demand responsiveness, social connection desire, empathy, and social continuity. Rejection reactivity measures the degree to which learners react to rejection and failure \cite{b64,b65}, both socially and cognitively. Demand responsiveness measures learner performance of conformity and compliance with instructions and expectations \cite{b12,b38,b60,b66,b67}. Social connection desire measures the degree to which learners actively seek out and desire help, social opportunities, and a sense of belonging from peers and faculty in the online university setting \cite{b61,b67,b68,b69}. The Empathy scale characterizes more collaborative \cite{b70} interactions \cite{b45} for learners with the ability to attune to the feelings and perspectives of others and respond with compassion \cite{b71} avoiding both overaccommodating and indifferent detachment \cite{b72}. Social continuity measures learners' levels of saturation in stable and enduring social relationships \cite{b73} outside the university setting, as well as their bandwidth in forming and maintaining new bonds \cite{b41,b74}. 
\subsection{Reasoning}
The Reasoning category is comprised of five cognitive scales essential to student success \cite{b75}: flexibility of thought, emotional-rational balance, balanced decisiveness, risk inclination, and engagement consistency. Flexibility of thought measures the degree to which learners can adapt \cite{b3,b45} their thinking and beliefs in response to new information or changing circumstances \cite{b38,b76} to solve problems.
Emotional-rational balance is defined by the range between the subjectivity of emotional thinking and the, at times, detached \cite{b45} banality of overly logical thinking contributing to a learner’s overall emotional intelligence \cite{b70,b77}. The balanced decisiveness scale assesses learners’ ability \cite{b79} to make timely and effective decisions in support of their academic goals \cite{b24, b80}. The risk inclination scale measures learner comfortability \cite{b81} with risk-taking \cite{b82}. Engagement consistency contrasts learners’ observable or perceived tendencies toward reliability \cite{b61} versus impulsivity \cite{b83, b84} in their academic performance \cite{b55}.
\subsection{Executive Function}
The Executive Function category is comprised of five scales related to learners' efficiency: time management ability, concentration, task approach, memory function, and systems thinking. The time management ability scale measures the effectiveness \cite{b5,b38} of learners’ internal capacity for organizing, planning, initiating, and following through on academic commitments \cite{b85}, differentiated from their observable engagement consistency. The concentration scale is defined as a learner’s range from hyperfocus to distractibility \cite{b38, b86}. The task approaches scales consider learners’ tendencies toward polychronicity (multitasking) versus monochronicity (single-tasking) \cite{b38,b76,b87,b88} and toward accessing repetitive or ritualistic behaviors in support of their learning \cite{b66} versus varied activities and approaches. The memory scale is defined by a learner’s working and short-term memory capacity \cite{b89} and its impact \cite{b90,b91}. The systems thinking scale measures learners' granularity of focus, whether they are conscientious of small details or more big-picture-oriented \cite{b92}.
\subsection{Core Skills}
The final category of GEM considers learners’ foundational efficacy \cite{b93} in core skills for academic success \cite{b94}, including familiarity with the field of study, literacy, verbal articulation and comprehension, technology, and the learners’ own self-awareness. The field of study scale measures learners’ efficacy with their subject matter at the start of their program and continually. The literacy scale measures learner confidence in reading and writing at the expected level of the established language \cite{b95}. The verbal articulation and comprehension scale measures a learners’ ability to express themselves through spoken word and understand spoken language effectively \cite{b96,b97}. The technology efficacy scale relates to a learner’s proficiency in and access to using varied and evolving digital tools and resources for academic purposes \cite{b98,b99,b100}. The self-awareness scale assesses learners’ understanding of their own strengths, weaknesses, and learning requirements \cite{b20}.

\section{Personas}
\begin{figure*}[ht] 
      \centering
  
    \pdftooltip{\includegraphics[width=0.8\linewidth]{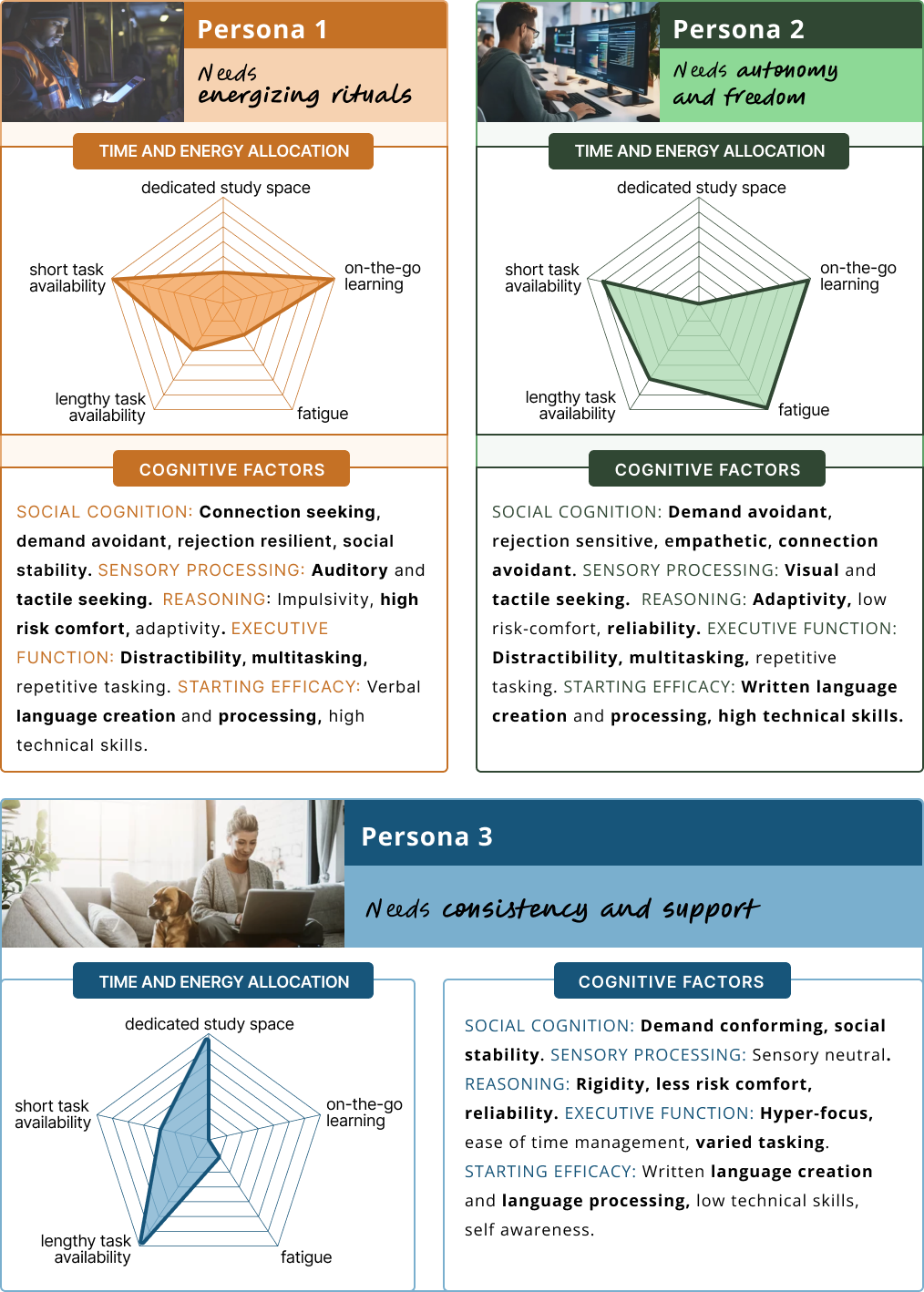}}{The infographic for Persona One shows an image of a young male looking at a mobile phone screen while riding a bus. Title: Persona 1. Subtitle: Needs Energizing Rituals. The Time and Energy Allocation is depicted using a five-point radial chart that, for Persona 1, indicates high availability for short tasks, medium availability for length tasks, medium fatigue levels, high levels of on-the-go learning, and medium availability of a dedicated study space. The Cognitive Factors associated with this persona include: Connection seeking, rejection resilience, risk comfort, multitasking, auditory and tactile seeking and verbal language skills. The infographic for Persona Two shows an image of a male learner in front of a computer screen. Title: Persona 2, subtitle: Needs Autonomy and Freedom. The Time and Energy Allocation is depicted using a five-point radial chart that, for Persona 2, indicates high availability for short tasks, medium availability for lengthy tasks, high fatigue levels, high levels of on-the-go learning, and very low availability of a dedicated study space. Cognitive factors include Demand-avoidant, rejection sensitive, empathetic, repetitive tasking, reliability, written language creation and technical skills. The infographic for Persona Three shows an image of a female working in a home environment. Title: Persona 3. Subtitle: Needs Consistency and Support. The Time and Energy Allocation is depicted using a five-point radial chart that for Persona 3 indicates low availability for short tasks, high availability for lengthy tasks, low fatigue levels, low levels of on-the go learning, and high availability of a dedicated study space. Cognitive factors include Demand conforming, sensory neutrality, rigidity, low-risk comfort, hyper-focus, low technical skills, and self-awareness.}
     \caption{Three sample personas: Persona 1 - "Energizing Rituals," Persona 2 - "Autonomy and Freedom," Persona 3 - "Consistency and Support"}
     
    \label{fig:personae}
\end{figure*}

In order to evaluate the breadth of GEM and illustrate the model through context, problems, and needs, this section introduces three sample learner profiles (see \hyperref[fig:personae]{Fig.~\ref{fig:personae}} after References). The authors opted not to reference clinically-defined neurodiversity in the personas, focusing instead on specific factors that influence learning experiences. This perspective acknowledges that neurodivergent individuals can often be indistinguishable from neurotypicals and seeks to avoid perpetuating stereotypes by not adhering to restrictive definitions of spectrum conditions.

The first learner profile, Persona 1, needs energizing rituals. They are tenacious and friendly. They use breaks and wait times throughout their daily activities for energizing learning activities that prime them to study deeply later at a designated workspace. They seek social interactions with their online school community, preferring spoken interactions and audio options for course engagement. Their impulsivity and distractibility may hinder effective time management, causing missed scheduled live events or risky decision-making. However, they recover from rejection or failure with notable resilience.

The second learner profile, Persona 2, values personal autonomy and freedom deeply and is highly sensory and rejection sensitive. In their quest for control, they may face challenges when required to follow specific directions or solve problems in a prescribed manner. They often structure their own study approach and schedule, for example, learning constantly in small chunks while on-the-go. They leverage their technical skills to customize their digital learning experience to suit their ongoing fluctuating needs, though their self-awareness may occasionally be insufficient for them to recognize those needs. Finding spoken and auditory communication challenging, they prefer anonymity in social situations or opt to avoid them altogether while also taking pride in their reliability, not wanting to let others down. However, this penchant for independence may lead to high burnout potential due to over-commitment and frequent multitasking.

The third learner profile, Persona 3, prefers the familiar. They rely on a prescribed structure and very clear directions for their learning, often seeking guidance from traditional instructors. They compartmentalize academic and personal relationships and seek social support outside formal academic settings. They typically react neutrally to sensory stimuli, and they are equally comfortable expressing themselves in various communication modalities. Strong time management skills and a stable, dedicated workspace enable them to stay focused on academic tasks for extended periods of time, though they may exhibit a lack of adaptability to changes in routine, environment, or learning platforms. However, they are typically aware of their limitations, such as technical skills, and seek assistance proactively when necessary.

\section{Decisions and Alternatives}
The three sample learner profiles characterize examples of the distinct and diverse momentary, fluctuating \cite{b58}, and static needs of learners in an asynchronous online competency-based learning environment. A four-step process is outlined to address these needs at scale and facilitate inclusive educational experiences: 1) identify learner behavioral patterns; 2) identify learner contexts when exhibiting the behaviors; 3) integrate a system of holistic learning support; 4) combine behavioral data and ongoing context data to surface just-in-time support tool recommendations for a learner. 
\subsection{Step 1: Identify learner behavioral patterns}
The initial step involves examining learner behaviors in the online platform, such as engagement (i.e., how frequently and for how long learners engaged with the learning platform and accuracy of formative and summative activities attempted). Educators will implement learning platform telemetry and instrumentation augmented by generative AI \cite{b23} to identify learner behavioral patterns and technical problems.
\subsection{Step 2: Explore learner context patterns} 
The second step explores the contextual underpinnings to learners’ behavior. Educators will introduce frequent in-platform checkpoints from the point of orientation and throughout the program of study. These checkpoints will include psychometrically-validated questions aligned to the GEM scales to interpret fluctuating learner mindsets and environments. The results will continually evaluate the effectiveness of GEM and the prioritization of the factors to be addressed, leading to prioritized technology enablement needs for students with neurodivergent traits that would otherwise be unsupported.\\

\subsection{Step 3: Integrate systems of holistic support }
Building a culture of inclusive thinking for student support and supportive technical tools involves both enabling supportive technical capabilities and reframing \cite{b71} faculty and staff mindsets to a trauma-informed lens \cite{b26}. Some possibilities initially identified through examination of the personas for technology support include customizable digital experiences, multi-modal collaboration, enhanced assessment experience, note-taking task scheduling, and AI-assisted co-regulation. 
Customizable digital experience enables the learner's ability to adjust key aspects of the learning platform, such as background and font color, font size and type, haptic settings, and read-aloud functions. In multi-modal collaboration, instructors engage with learners using learners' preferred communication modes, including text, email, voice, or video. These same options are available for peer connections and collaboration via an online staff-moderated virtual community space. The virtual community space supports both course and program spaces as well as student extracurricular or special interest support spaces. Enhanced assessment considers rejection reactivity within formative and summative assessment feedback (including trauma-informed proctoring) and enables various modes of formative activities for access across devices. Note-taking enables annotation and commenting within the learning materials, along with a summary area where those notes can be viewed, searched using keywords, or exported and shared with peers, faculty, and other apps on the learner's device as requested. Task scheduling enables a checklist view of all tasks for a course or program, a calendar view with scheduling and exporting of tasks and events, and digital reminders and alerts for tasks and activities suggested by the platform or created by the learner. AI-assisted co-regulation refers to modeling emotion identification and productive coping to guide learners in developing self-regulation skills, such as through the use of a sensory diet \cite{b71}.

\subsection{Step 4: Implement guiding empowerment}
In the final step, behavioral data and ongoing contextual checkpoint data are combined to identify and surface the data-informed, tailored support tool recommendations to address learners' evolving needs and challenges. For instance, Persona 1 may benefit from customizable task scheduling and alerts, read-aloud functions, and quick haptics-enabled formative practice activities accessible on mobile devices. Persona 2 may benefit from modifiable content background colors, trauma-informed proctoring, and text-based peer and instructor engagement and course materials. Lastly, Persona 3 may benefit from structured task scheduling, technical support, and a higher degree of consistency across their program. 

\subsection{Generative AI} 
Generative AI should be considered as an interwoven method to guide \cite{b23,b101} learners by identifying and recommending highlighted learning and support tools as the continuous learning capabilities of these systems will sustain scalability. For instance, the checkpoints in Step 2 could trigger a notification for an instructor to make contact with the student using a specific mode and tone or trigger direct intervention from the platform itself in the form of automatic alerts, personalized content and experience, and other guidance.

\section{Limitations}
 Learners' data security, human quality checks of AI recommendations and the specifics of the learner support system are all facets of this approach that must be carefully considered and defined in implementation. Learners must opt-in to their data being tracked and receiving data-adjusted experience recommendations. The format of the frequent checkpoints in the platform must not distract or impose on learners. The platform experience (e.g. the activities recommended, the course material and assessment modes, the course community activities highlighted, and the communication modes by instructors) should adapt dynamically based on learners' ongoing responses rather than categorizing them into one static persona or profile. GEM is an exploratory conceptual model that must be tested in practice empirically before the impact on learner achievement can be determined.
 
\section{Conclusion}
This article has offered researchers and online education providers an approach to a deliberate, active commitment to equity within a neurodiverse learner population that has historically excluded and under-supported neurodivergent learners and those with situational limitations. 
The authors advocate for learners’ dignity, informed agency, and accommodations in the ways they study and live and propose that addressing the contextual needs computing and engineering learners face while studying in an online, asynchronous, autonomous learning environment will improve the achievement of not only officially diagnosed neurodivergent learners but also those with undiagnosed cognitive challenges or situational limitations. Furthermore, this article has suggested a process for identifying the various fluctuating needs these populations may face and for building systems of equitable learning support and a culture of inclusivity and data-informed personalization.   

\section{Disclosure}
The first author discloses as a diagnosed autistic person and a member of the neurodiversity movement, an identity that affects the interpretation of the secondary research compiled and the perspectives expressed in this article.

\bibliographystyle{plain} 

\begin{thebibliography}{100}

\bibitem{b27}
A.~Al-Azawei, F.~Serenelli, and K.~Lundqvist.
\newblock Universal design for learning (udl): A content analysis of peer-reviewed journals from 2012 to 2015.
\newblock {\em Journal of Scholarship of Teaching and Learning}, 16(3):39--56, 2016.

\bibitem{b91}
T.~P. Alloway.
\newblock A comparison of working memory profiles in children with adhd and dcd.
\newblock {\em Child Neuropsychology}, 17(5):483--494, 2011.

\bibitem{b50}
M.~Anghel.
\newblock The influence of sensory systems in motor development to the preschool child.
\newblock {\em Bulletin of the Transilvania University of Brașov. Series IX: Sciences of Human Kinetics}, 12(61)(1):189--194, Jun. 2019.

\bibitem{b89}
R.~E. Avery, L.~D. Smillie, and J.~W.~De Fockert.
\newblock The role of working memory in achievement goal pursuit.
\newblock {\em Acta Psychologica}, 144(2):361--372, Oct. 2013.

\bibitem{b93}
A.~Bandura.
\newblock Perceived self-efficacy in the exercise of personal agency.
\newblock {\em Journal of Applied Sport Psychology}, 2(2):128--163, 1990.

\bibitem{b57}
T.~Bar-Shalita, J.-J. Vatine, and S.~Parush.
\newblock Sensory modulation disorder: A risk factor for participation in daily life activities.
\newblock {\em Developmental Medicine and Child Neurology}, 50(12):932--937, 2008.

\bibitem{b29}
L.~Barnard, J.~S. Yi, J.~A. Jacko, and A.~Sears.
\newblock Capturing the effects of context on human performance in mobile computing systems.
\newblock {\em Personal and Ubiquitous Computing}, 11:81--96, 2007.

\bibitem{b72}
J.~Bay.
\newblock Fostering diversity, equity, and inclusion in the technical and professional communication service course.
\newblock {\em IEEE Transactions on Professional Communication}, 65(1):213--225, Mar. 2022.

\bibitem{b76}
S.~Braem and T.~Egner.
\newblock Getting a grip on cognitive flexibility.
\newblock {\em Current Directions in Psychological Science}, 27(6):470--476, 2018.

\bibitem{b37}
L.~M. Carrier, L.~D. Rosen, N.~A. Cheever, and A.~F. Lim.
\newblock Causes, effects, and practicalities of everyday multitasking.
\newblock {\em Developmental Review}, 35:64--78, 2015.

\bibitem{b87}
L.~M. Carrier, L.~D. Rosen, N.~A. Cheever, and A.~F. Lim.
\newblock Causes, effects, and practicalities of everyday multitasking.
\newblock {\em Developmental Review}, 35:64--78, 2015.

\bibitem{b11}
CAST.
\newblock Universal design for learning guidelines website, 2018.
\newblock Accessed: Dec. 10, 2023.

\bibitem{b14}
M.~D.~B. Castro and G.~M. Tumibay.
\newblock A literature review: Efficacy of online learning courses for higher education institutions using meta-analysis.
\newblock {\em Education and Information Technologies}, 26(2):1367--1385, Nov. 2019.

\bibitem{b58}
T.~Champagne, J.~Koomar, and L.~Olson.
\newblock Sensory processing evaluation and intervention in mental health.
\newblock {\em OT Practice}, 15(5):CE1--CE7, 2010.

\bibitem{b8}
M.~Chrysochoou, A.~E. Zaghi, and C.~M. Syharat.
\newblock Reframing neurodiversity in engineering education.
\newblock {\em Frontiers in Education}, 7, Nov. 2022.

\bibitem{b43}
M.~Clince, L.~Connolly, and C.~Nolan.
\newblock Comparing and exploring the sensory processing patterns of higher education students with attention deficit hyperactivity disorder and autism spectrum disorder.
\newblock {\em The American Journal of Occupational Therapy}, 70(2):7002250010p1--7002250010p9, 2016.

\bibitem{b90}
N.~Cowan.
\newblock What are the differences between long-term, short-term, and working memory?
\newblock {\em Progress in Brain Research}, 169:323--338, 2008.

\bibitem{b3}
B.~E. Cox, J.~Edelstein, B.~Brogdon, and A.~Roy.
\newblock Navigating challenges to facilitate success for college students with autism.
\newblock {\em Journal of Higher Education}, 92(2):252--278, Aug. 2020.

\bibitem{b5}
B.~E. Cox, J.~Edelstein, B.~Brogdon, and A.~Roy.
\newblock Navigating challenges to facilitate success for college students with autism.
\newblock {\em Journal of Higher Education}, 92(2):252--278, Aug. 2020.

\bibitem{b1}
A.~R. Dallman, K.~L. Williams, and L.~Villa.
\newblock Neurodiversity-affirming practices are a moral imperative for occupational therapy.
\newblock {\em Open Journal of Occupational Therapy}, 10(2):1--9, Apr. 2022.

\bibitem{b39}
E.~E. Dean, L.~Little, S.~Tomchek, and W.~Dunn.
\newblock Sensory processing in the general population: Adaptability, resiliency, and challenging behavior.
\newblock {\em The American Journal of Occupational Therapy}, 72(1):7201195060p1--7201195060p8, 2018.

\bibitem{b100}
H.~Deng.
\newblock Emerging patterns and trends in utilizing electronic resources in a higher education environment: An empirical analysis.
\newblock {\em New Library World}, 111(3/4):87--103, 2010.

\bibitem{b92}
M.~A. Dolansky, S.~M. Moore, P.~A. Palmieri, and M.~K. Singh.
\newblock Development and validation of the systems thinking scale.
\newblock {\em Journal of General Internal Medicine}, 35(8):2314--2320, 2020.

\bibitem{b83}
A.~L. Duckworth and M.~E.~P. Seligman.
\newblock Self-discipline outdoes iq in predicting academic performance of adolescents.
\newblock {\em Psychological Science}, 16(12):939--944, 2005.

\bibitem{b49}
W.~Dunn, C.~Brown, A.~Breitmeyer, and A.~Salwei.
\newblock Corrigendum: Construct validity of the sensory profile interoception scale: Measuring sensory processing in everyday life.
\newblock {\em Frontiers in Psychology}, 13, 2022.

\bibitem{b44}
W.~Dunn, C.~Brown, and A.~McGuigan.
\newblock The ecology of human performance: A framework for considering the effect of context.
\newblock {\em The American Journal of Occupational Therapy}, 48(7):595--607, 1994.

\bibitem{b2}
P.~Dwyer.
\newblock The neurodiversity approach(es): What are they and what do they mean for researchers?
\newblock {\em Human Development}, 66(2):73--92, Jan. 2022.

\bibitem{b51}
M.~O. Ernst.
\newblock The 'puzzle' of sensory perception: putting together multisensory information.
\newblock In {\em Proc. 7th Int. Conf. on Multimodal interfaces}, pages 1--1, 2005.

\bibitem{b19}
J.~L. Rivera~Munoz et~al.
\newblock Systematic review of adaptive learning technology for learning in higher education.
\newblock {\em Eurasian Journal of Educational Research}, 98:221--233, 2022.

\bibitem{b32}
M.~E. King-Sears et~al.
\newblock Achievement of learners receiving udl instruction: A meta-analysis.
\newblock {\em Teaching and Teacher Education}, 122:103956, 2023.

\bibitem{b15}
S.~Palvia et~al.
\newblock Online education: Worldwide status, challenges, trends, and implications.
\newblock {\em Journal of Global Information Technology Management}, 21(4):233--241, Oct. 2018.

\bibitem{b95}
A.~Fernández-Villardón, R.~Valls-Carol, P.~Melgar Alcantud, and I.~Tellado.
\newblock Enhancing literacy and communicative skills of students with disabilities in special schools through dialogic literary gatherings.
\newblock {\em Frontiers in Psychology}, 12, 2021.

\bibitem{b77}
M.~Fiori, S.~Udayar, and A.~V. Maillefer.
\newblock Emotion information processing as a new component of emotional intelligence: Theoretical framework and empirical evidence.
\newblock {\em European Journal of Personality}, 36(2):245--264, 2022.

\bibitem{b41}
L.~D. Goegan and L.~M. Daniels.
\newblock Academic success for students in postsecondary education: The role of student characteristics and integration.
\newblock {\em Journal of College Student Retention: Research, Theory and Practice}, 23(3):659--685, 2021.

\bibitem{b64}
S.~D. Gunawardena, P.~Devine, I.~Beaumont, L.~P. Garden, E.~Murphy-Hill, and K.~Blincoe.
\newblock Destructive criticism in software code review impacts inclusion.
\newblock In {\em Proc. ACM on Human-Computer Interaction}, volume~6, pages 1--29, 2022.

\bibitem{b20}
L.~G. Hamilton and S.~Petty.
\newblock Compassionate pedagogy for neurodiversity in higher education: A conceptual analysis.
\newblock {\em Frontiers in Psychology}, 14, 2023.

\bibitem{b25}
S.~L. Henry, S.~Abou-Zahra, and J.~Brewer.
\newblock The role of accessibility in a universal web.
\newblock In {\em Proceedings of the 11th Web for All Conference}, pages 1--4, 2014.

\bibitem{b23}
A.~Iku-Silan, G.-J. Hwang, and C.-H. Chen.
\newblock Decision-guided chatbots and cognitive styles in interdisciplinary learning.
\newblock {\em Computers and Education}, 201, 2023.

\bibitem{b62}
R.~J. Jagers, D.~Rivas-Drake, and B.~Williams.
\newblock Transformative social and emotional learning (sel): Toward sel in service of educational equity and excellence.
\newblock {\em Educational Psychologist}, 54(3):162--184, 2019.

\bibitem{b101}
E.~Jenks, F.~Selman, M.~Harmens, S.~Boon, T.~Tran, H.~Hobson, S.~Eagle, and F.~Sedgewick.
\newblock Teaching higher education staff to understand and support autistic students: evaluation of a novel training program.
\newblock {\em Frontiers in Psychiatry}, 14, 2023.

\bibitem{b38}
N.~Kenny and A.~Doyle.
\newblock I have always lived desperate and vulnerable on the edge of helplessness and collapse…: A phenomenological exploration of the lived experience of adults experiencing pathological demand avoidance, 2023.

\bibitem{b45}
N.~Kholifah and B.~Astuti.
\newblock The correlation between peer social interaction with resilience in student.
\newblock In {\em Proceedings of the 5th International Conference on Learning Innovation and Quality Education}, pages 1--5, 2021.

\bibitem{b40}
M.~Kinnealey, K.~P. Koenig, and S.~Smith.
\newblock Relationships between sensory modulation and social supports and health-related quality of life.
\newblock {\em The American Journal of Occupational Therapy}, 65(3):320--327, 2011.

\bibitem{b17}
C.~C. Kulik, J.~A. Kulik, and R.~L. Bangert-Drowns.
\newblock Effectiveness of mastery learning programs: A meta-analysis.
\newblock {\em Review of Educational Research}, 60(2):265--299, Jun. 1990.

\bibitem{b82}
J.~Leota, K.~Nash, and I.~McGregor.
\newblock Reactive risk-taking: anxiety regulation via approach motivation increases risk-taking behavior.
\newblock {\em Personality and Social Psychology Bulletin}, 49(1):81--96, 2023.

\bibitem{b70}
M.~P. Lillis.
\newblock Faculty emotional intelligence and student-faculty interactions: Implications for student retention.
\newblock {\em Journal of College Student Retention: Research, Theory \& Practice}, 13(2):155--178, 2011.

\bibitem{b94}
E.~A. Linnenbrink and P.~R. Pintrich.
\newblock The role of self-efficacy beliefs in student engagement and learning in the classroom.
\newblock {\em Reading and Writing Quarterly}, 19(2):119--137, 2003.

\bibitem{b85}
C.~MacCann, G.~J. Fogarty, and R.~D. Roberts.
\newblock Strategies for success in education: Time management is more important for part-time than full-time community college students.
\newblock {\em Learning and Individual Differences}, 22(5):618--623, 2012.

\bibitem{b54}
T.~Machingura, G.~Kaur, C.~Lloyd, S.~Mickan, D.~Shum, E.~Rathbone, and H.~Green.
\newblock An exploration of sensory processing patterns and their association with demographic factors in healthy adults.
\newblock {\em Irish Journal of Occupational Therapy}, 48(1):3--16, 2020.

\bibitem{b75}
F.~Martin, V.~P. Dennen, and C.~J. Bonk.
\newblock A synthesis of systematic review research on emerging learning environments and technologies.
\newblock {\em Educational Technology Research and Development}, 68:1613--1633, 2020.

\bibitem{b97}
S.~McWeeny and E.~S. Norton.
\newblock Auditory processing and reading disability: A systematic review and meta-analysis.
\newblock {\em Scientific Studies of Reading}, 28(2):167--189, 2024.

\bibitem{b61}
N.~B. Mendoza and R.~B. King.
\newblock The social contagion of student engagement in school.
\newblock {\em School Psychology International}, 41(5):454--474, 2020.

\bibitem{b34}
L.~J. Miller, S.~A. Schoen, S.~Mulligan, and J.~Sullivan.
\newblock Identification of sensory processing and integration symptom clusters: A preliminary study.
\newblock {\em Occupational Therapy International}, 2017.

\bibitem{b53}
L.~J. Miller, S.~A. Schoen, S.~Mulligan, and J.~Sullivan.
\newblock Identification of sensory processing and integration symptom clusters: A preliminary study.
\newblock {\em Occupational Therapy International}, 2017.

\bibitem{b73}
S.~Mishra.
\newblock Social networks, social capital, social support and academic success in higher education: A systematic review with a special focus on ‘underrepresented’ students.
\newblock {\em Educational Research Review}, 29, 2020.

\bibitem{b12}
A.~Moore.
\newblock Pathological demand avoidance: What and who are being pathologised and in whose interests?
\newblock {\em Global Studies of Childhood}, 10(1):39--52, 2020.

\bibitem{b31}
J.~W. Moore.
\newblock What is the sense of agency and why does it matter?
\newblock {\em Frontiers in Psychology}, 7, 2016.

\bibitem{b24}
A.~Moriña.
\newblock When what is unseen does not exist: disclosure, barriers and supports for students with invisible disabilities in higher education.
\newblock {\em Disability and Society}, pages 1--19, 2022.

\bibitem{b13}
D.~Murray, D.~Milton, J.~Green, and J.~Bervoets.
\newblock The human spectrum: A phenomenological enquiry within neurodiversity.
\newblock {\em Psychopathology}, 56(3):220--230, 2023.

\bibitem{b30}
L.~A. Newman, J.~W. Madaus, A.~R. Lalor, and H.~S. Javitz.
\newblock Support receipt: Effect on postsecondary success of students with learning disabilities.
\newblock {\em Career Development and Transition for Exceptional Individuals}, 42(1):6--16, 2019.

\bibitem{b21}
T.~R. Nodine.
\newblock How did we get here? a brief history of competency-based higher education in the united states.
\newblock {\em Journal of Competency-Based Education}, 1(1):5--11, Apr. 2016.

\bibitem{b55}
C.~Nolan, J.~K. Doyle, K.~Lewis, and D.~Treanor.
\newblock Disabled students' perception of the sensory aspects of the learning and social environments within one higher education institution.
\newblock {\em British Journal of Occupational Therapy}, 86(5):367--375, 2023.

\bibitem{b60}
M.~Nwabuoku.
\newblock Surviving distance learning as an adult learner in higher education.
\newblock {\em Adult Learning}, 31(4):185--187, 2020.

\bibitem{b66}
E.~ONions, E.~Viding, C.~U. Greven, A.~Ronald, and F.~Happe.
\newblock Pathological demand avoidance: exploring the behavioural profile.
\newblock {\em Autism}, 18(5):538--544, 2014.

\bibitem{b16}
F.~Ouyang, L.~Zheng, and P.~Jiao.
\newblock Artificial intelligence in online higher education: A systematic review of empirical research from 2011 to 2020.
\newblock {\em Education and Information Technologies}, 27(6):7893--7925, Feb. 2022.

\bibitem{b22}
M.~Papadatou-Pastou, A.~K. Touloumakos, and C.~Koutouveli et~al.
\newblock The learning styles neuromyth: when the same term means different things to different teachers.
\newblock {\em European Journal of Psychology of Education}, 36:511--531, 2021.

\bibitem{b79}
A.~L. Patalano and S.~M. Wengrovitz.
\newblock Indecisiveness and response to risk in deciding when to decide.
\newblock {\em Journal of Behavioral Decision Making}, 20(4):405--424, 2007.

\bibitem{b47}
S.~A. Powers and M.~W. Scerbo.
\newblock Examining the effect of interruptions at different breakpoints and frequencies within a task.
\newblock {\em Human Factors}, 65(1):22--36, 2023.

\bibitem{b42}
V.~L. Reese and R.~Dunn.
\newblock Learning-style preferences of a diverse freshmen population in a large, private, metropolitan university by gender and gpa.
\newblock {\em Journal of College Student Retention: Research, Theory and Practice}, 9(1):95--112, 2007.

\bibitem{b102}
K.~A. Robert.
\newblock {\em The Vigor of Creative Materialism: Making the Hidden Stories of Underrepresented Engineering Students Visible}.
\newblock PhD thesis, University of Denver, Denver, CO, USA, 2023.

\bibitem{b46}
J.~Roksa and P.~Kinsley.
\newblock The role of family support in facilitating academic success of low-income students.
\newblock {\em Research in Higher Education}, 60:415--436, 2019.

\bibitem{b33}
S.~R. Ross.
\newblock Supporting your neurodiverse student population with the universal design for learning (udl) framework.
\newblock In {\em 2019 IEEE Frontiers in Education Conference (FIE)}, 2019.

\bibitem{b96}
S.~C. Schwering and M.~C. MacDonald.
\newblock Verbal working memory as emergent from language comprehension and production.
\newblock {\em Frontiers in Human Neuroscience}, 14, 2020.

\bibitem{b28}
S.~Seok, B.~DaCosta, and R.~Hodges et~al.
\newblock A systematic review of empirically based universal design for learning: Implementation and effectiveness of universal design in education for students with and without disabilities at the postsecondary level.
\newblock {\em Open Journal of Social Sciences}, 6(05):171, 2018.

\bibitem{b99}
I.~M. Shaikh, A.~Alsharief, H.~Amin, K.~Noordin, and J.~Shaikh.
\newblock Inspiring academic confidence in university students: perceived digital experience as a source of self-efficacy.
\newblock {\em On the Horizon: The International Journal of Learning Futures}, 31(2):110--122, 2023.

\bibitem{b52}
S.~Shimojo and L.~Shams.
\newblock Sensory modalities are not separate modalities: plasticity and interactions.
\newblock {\em Nature Reviews Neuroscience}, 2(7):505--513, Jul. 2001.

\bibitem{b10}
S.~Shipman and V.~C. Shipman.
\newblock Chapter 7: Cognitive styles: Some conceptual, methodological, and applied issues.
\newblock {\em Review of Research in Education}, 12(1):229--291, 1985.

\bibitem{b81}
V.~K. Singh, R.~Goyal, and S.~Wu.
\newblock Riskalyzer: Inferring individual risk-taking propensity using phone metadata.
\newblock In {\em Proc. of the ACM on Interactive, Mobile, Wearable and Ubiquitous Technologies}, volume~2, pages 1--21, 2018.

\bibitem{b36}
A.~Strantz.
\newblock Using web standards to design accessible data visualizations in professional communication.
\newblock {\em IEEE Transactions on Professional Communication}, 64(3):288--301, Sept. 2021.

\bibitem{b67}
S.~M. Suldo, R.~F. Dedrick, E.~Shaunessy-Dedrick, S.~A. Fefer, and J.~Ferron.
\newblock Development and initial validation of the coping with academic demands scale: How students in accelerated high school curricula cope with school-related stressors.
\newblock {\em Journal of Psychoeducational Assessment}, 33(4):357--374, 2015.

\bibitem{b69}
N.~Sun and M.~B. Rosson.
\newblock Connection enablers in online learning community: From informative online personae to meaningful social space.
\newblock In {\em Companion of the 2017 ACM Conference on Computer Supported Cooperative Work and Social Computing}, pages 315--318, 2017.

\bibitem{b68}
N.~Sun, X.~Wang, and M.~B. Rosson.
\newblock How do distance learners connect?
\newblock In {\em Proc. of the 2019 CHI Conference on Human Factors in Computing Systems}, pages 1--12, 2019.

\bibitem{b4}
C.~M. Syharat, A.~Hain, A.~E. Zaghi, R.~Gabriel, and C.~G.~P. Berdanier.
\newblock Experiences of neurodivergent students in graduate stem programs.
\newblock {\em Frontiers in Psychology}, 14, Jun. 2023.

\bibitem{b63}
A.~L.~L. Tang, C.~Walker-Gleaves, and J.~Rattray.
\newblock An exploratory study of university teachers’ conceptions and articulation of care amidst online teaching.
\newblock {\em Pastoral Care in Education}, pages 1--21, 2022.

\bibitem{b65}
V.~Y.~K. Tao, Y.~Li, and A.~M.~S. Wu.
\newblock Incremental intelligence mindset, fear of failure, and academic coping.
\newblock {\em Journal of Pacific Rim Psychology}, 16, 2022.

\bibitem{b74}
S.~E. Taylor, J.~A. Scepansky, J.~W. Lounsbury, and L.~W. Gibson.
\newblock Broad and narrow personality traits of women’s college students in relation to early departure from college.
\newblock {\em Journal of College Student Retention: Research, Theory \& Practice}, 11(4):483--497, 2010.

\bibitem{b59}
N.~Tottenham, T.~A. Hare, and B.~J. Casey.
\newblock Behavioral assessment of emotion discrimination, emotion regulation, and cognitive control in childhood, adolescence, and adulthood.
\newblock {\em Frontiers in Psychology}, 2, 2011.

\bibitem{b48}
S.~Tremblay, F.~Vachon, D.~Lafond, and C.~Kramer.
\newblock Dealing with task interruptions in complex dynamic environments: Are two heads better than one?
\newblock {\em Human Factors}, 54(1):70--83, 2012.

\bibitem{b9}
J.~Treviranus.
\newblock Authoring tools.
\newblock In {\em Web Accessibility: A Foundation for Research}, pages 127--138. Springer, 2008.

\bibitem{b7}
M.~Tsugawa, B.~Webster, S.~Solanki, A.~Cuellar, and C.~Spence.
\newblock Examination of ableist educational systems and structures that limit access to engineering education through narratives.
\newblock In {\em Proceedings of ASEE Annual Conference}, Feb. 2024.

\bibitem{b6}
Y.~Turjeman-Levi and A.~N. Kluger.
\newblock Sensory-processing sensitivity versus the sensory-processing theory: Convergence and divergence.
\newblock {\em Frontiers in Psychology}, 13, Dec. 2022.

\bibitem{b56}
Y.~Turjeman-Levi and A.~N. Kluger.
\newblock Sensory-processing sensitivity versus the sensory-processing theory: Convergence and divergence.
\newblock {\em Frontiers in Psychology}, 13, 2022.

\bibitem{b26}
A.~S. Venet.
\newblock {\em Equity-Centered Trauma-Informed Education}.
\newblock Routledge, New York, NY, USA, 2023.

\bibitem{b98}
C.-H. Wang, D.~M. Shannon, and M.~E. Ross.
\newblock Students’ characteristics, self-regulated learning, technology self-efficacy, and course outcomes in online learning.
\newblock {\em Distance Education}, 34(3):302--323, 2013.

\bibitem{b35}
D.~Wang, S.~Casares, K.~Eilers, S.~Hitchcock, R.~Iverson, E.~Lahn, M.~Loux, C.~Schnetzer, and L.~A. Frey-Law.
\newblock Assessing multisensory sensitivity across scales: Using the resulting core factors to create the multisensory amplification scale.
\newblock {\em The Journal of Pain}, 23(2):276--288, 2022.

\bibitem{b86}
K.~D. Wang, J.~McCool, and C.~Wieman.
\newblock Exploring the learning experiences of neurodivergent college students in stem courses.
\newblock {\em Journal of Research in Special Educational Needs}, 00:1--14, 2024.

\bibitem{b80}
W.~J. Wellington, D.~B. Hutchinson, and A.~J. Faria.
\newblock Measuring the impact of a marketing simulation game: Experience on perceived indecisiveness.
\newblock {\em Simulation \& Gaming}, 48(1):56--80, 2017.

\bibitem{b18}
M.~Winget and A.~M. Persky.
\newblock A practical review of mastery learning.
\newblock {\em American Journal of Pharmaceutical Education}, 86(10):ajpe8906, Dec. 2022.

\bibitem{b88}
T.~Wu, J.~Gao, and L.~Wang.
\newblock Exploring links between polychronicity and job performance from the person–environment fit perspective—the mediating role of well-being.
\newblock {\em International Journal of Environmental Research and Public Health}, 17(10), 2020.

\bibitem{b84}
J.~Zhou and S.~Bhat.
\newblock Modeling consistency using engagement patterns in online courses.
\newblock In {\em LAK21: 11th International Learning Analytics and Knowledge Conference}, pages 226--236, 2021.

\bibitem{b71}
A.~Zolyomi and J.~Snyder.
\newblock Social-emotional-sensory design map for affective computing informed by neurodivergent experiences.
\newblock In {\em Proc. ACM Hum.-Comput. Interact.}, volume~5, April 2021.

\end{thebibliography}

\clearpage

\end{document}